\documentclass[11pt,a4paper]{article}
\usepackage[pdftex]{graphicx}
\usepackage[hyperref]{acl2020}
\usepackage{times}
\usepackage{latexsym}

\usepackage{CJKutf8}
\usepackage{rotating}
\newcommand{\argmax}{\mathop{\rm arg~max}\limits}

\newcommand{\wait}{\text{\textless wait\textgreater} }
\newcommand{\waits}{\textit{w} }
\newcommand{\cjk}[1]{\begin{CJK}{UTF8}{min}{#1}\end{CJK}}

\usepackage{breakurl}
\usepackage{xcolor}
\definecolor{darkblue}{rgb}{0, 0, 0.5}
\hypersetup{colorlinks=true,citecolor=darkblue, linkcolor=darkblue, urlcolor=darkblue}

\usepackage{color} 
\usepackage{ulem} 

\usepackage{amsmath}
\usepackage{amssymb}
\usepackage{amsfonts}
\usepackage{bm}
\usepackage{here}
\usepackage{multirow}
\usepackage{comment}

\aclfinalcopy 

\title{Simultaneous Neural Machine Translation \\using Connectionist Temporal Classification}

\author{Katsuki Chousa, Katsuhito Sudoh, Satoshi Nakamura \\
  Nara Institute of Science and Technology
\\}

\date{}

\begin{document}
\maketitle
\begin{abstract}
Simultaneous machine translation is a variant of machine translation that starts the translation process before the end of an input.
This task faces a trade-off between translation accuracy and latency.
We have to determine when we start the translation for observed inputs so far, to achieve good practical performance.
In this work, we propose a neural machine translation method to determine this timing in an adaptive manner.
The proposed method introduces a special token ‘\wait ’, which is generated when the translation model chooses to read the next input token instead of generating an output token.
It also introduces an objective function to handle the ambiguity in wait timings that can be optimized using an algorithm called Connectionist Temporal Classification (CTC).
The use of CTC enables the optimization to consider all possible output sequences including ‘\wait ' that are equivalent to the reference translations and to choose the best one adaptively.
We apply the proposed method into simultaneous translation from English to Japanese and investigate its performance and remaining problems.
\end{abstract}

\section{Introduction}
Simultaneous translation is a translation task where the translation process starts before the end of an input.
It helps real-time spoken language communications such as human conversations and public talks.
A usual machine translation system works in the sentence level and starts its translation process after it reads the end of a sentence.
It would not be appropriate for spoken languages due to roughly two issues: (1) sentence boundaries are not clear and (2) a large latency occurs for a long input.

Previous studies tackled this problem by an incremental process,
in order to reduce the translation latency for a given input.
\newcite{fujita13interspeech} proposed a phrase-based approach to the simultaneous translation
based on phrasal reordering probabilities.
\newcite{oda-etal-2015-syntax} proposed a syntax-based method to determine when to start translation of observed inputs.
Such an approach faces a trade-off between speed and accuracy; reducing the translation latency using very limited context information also causes the loss in the translation accuracy.
This becomes more serious especially in a syntactically-distant language pair such as English and Japanese,
where we sometimes have to wait a latter part of a source language to determine the corresponding former part in a target language.

Recent neural machine translation (NMT) studies tried an incremental processing for the simultaneous translation.
\newcite{gu2017learning} proposed a reinforcement learning approach to determine \textit{when to translate}
based on two different actions: READ to take one input token and WRITE to generate one output token.
While they reported some latency reduction without the loss of translation accuracy,
the NMT model itself is trained independently from this incremental manner and is not fully optimized for simultaneous translation.
\newcite{ma2018stacl} proposed a very simple incremental method called \textit{Wait-k},
where the decoder starts to generate output tokens after the encoder reads \textit{k} tokens and then works token-by-token.
Here, some required inputs may not be observed by the encoder;
however, the decoder has to predict the next output token even in that case.
This approach enables a simple end-to-end simultaneous NMT with implicit anticipation of unobserved inputs. 
It showed high translation accuracy with small latency on some common English-to-German and Chinese-to-English datasets.
The latency hyperparameter \textit{k} can be used to control the speed-accuracy trade-off,
but it has to be large enough for a distant language pair like English-Japanese.
We observed a problem in translating a phrase longer than \textit{k} tokens in our pilot study on English-to-Japanese translation.

In this work, we propose a novel incremental NMT method that uses a special token \wait in the target language
which is generated when the translation model chooses to read the next input token instead of generating an output token.
The proposed method uses Connectionist Temporal Classification (CTC) \cite{graves2006connectionist}
to handle ambiguities in possible positions inserting \wait in the training time.
CTC is applied to sequential model training such as automatic speech recognition,
where we have a reference word sequence but do not have the corresponding segmentation or alignment in an acoustic signal.
We conduct experiments in English-to-Japanese simultaneous translation with the proposed and baseline methods
and show the proposed method achieves a good translation performance with relatively small latency.
The proposed method can determine when to wait or translate in an adaptive manner and is useful in simultaneous translation tasks.


\section{Simultaneous machine translation by \textit{Wait-k} model}
First, we review a general NMT model following the formulation by \citet{luong-pham-manning:2015:EMNLP} and the ``Wait-k" model \cite{ma2018stacl} that is the baseline model for simultaneous NMT.

Given a source sentence $X$ and a target sentence $Y$ as follows:
\begin{align*}
    X=\{\textbf{x}_1, \textbf{x}_2, ..., \textbf{x}_I\}, \\
    Y=\{\textbf{y}_1, \textbf{y}_2, ..., \textbf{y}_J\},
\end{align*}
where $\textbf{x}_i \in \mathbb{R}^{S \times 1}$ is a one-hot vector of the i-th input word, 
$I$ is the length of the input sentence $X$, 
$\textbf{y}_i \in \mathbb{R}^{T \times 1}$ is a one-hot vector of the i-th output word, 
and $J$ is the length of the output sentence $Y$.

The problem of translation from the source to the target language can be solved by finding the best target language sentence $\hat{Y}$ that maximizes the conditional probability
\begin{equation}
    \hat{Y} = \argmax_Y p ( Y | X ) .
\end{equation}
In general NMT manner, the conditional probability is decomposed by the product of conditional generation probabilities of $\textbf{y}_{j}$ given the source sentence $X$ and preceding target words $\textbf{y}_{<j}$:
\begin{equation}
    p (Y | X) = \prod_{j=1}^{J} p_\theta (\textbf{y}_j | \textbf{y}_{<j}, X), \label{eq:general_word_prob}
\end{equation}
where $\textbf{y}_{<j}$ represents the target words up to position $j$, and $\theta$ indicates the model parameters.
In contrast, the model for simultaneous translation has to output translated words given only prefix words of the source sentence.
Therefore, the conditional probability is decomposed as follows:
\begin{equation}
    p (Y | X) = \prod_{j=1}^{J} p_\theta (\textbf{y}_j | \textbf{y}_{<j}, \textbf{x}_{<g(j)}), \label{eq:simultaneous_word_prob}
\end{equation}
where $\textbf{x}_{<g(j)}$ are the target words up to position $g(j)$ and $g(j)$ represents the number of encoded source tokens when the model outputs $j$ words.
In the ``Wait-k" model, $g(j)$ is defined as follows:
\begin{equation}
    g(j) = \begin{cases}
        k + j - 1 & (j < I - k) \\
        I & (\mathrm{otherwise})
    \end{cases}
\end{equation}
Here, $k$ is the hyperparameter which indicates the target sentence generation is $k$ tokens behind the source sentence input and it takes a constant value in the ``Wait-k" model.

The model is composed of an encoder (\S\ref{seq:encoder}) and a decoder with the attention mechanism (\S\ref{seq:attention_and_decoder}) that are both implemented using recurrent neural networks (RNNs); 
the encoder converts source words into a sequence of vectors, and the decoder generates target language words one-by-one with the attention mechanism based on the conditional probability shown in the equation \ref{eq:general_word_prob} and \ref{eq:simultaneous_word_prob}. 
The details are described below.

\subsection{Encoder}
\label{seq:encoder}
The encoder takes a sequence of a source sentence $X$ as inputs and returns forward hidden vectors $\overrightarrow{\textbf{h}_i}(1 \leq i \leq I)$ of the forward RNNs:
\begin{equation}
    \overrightarrow{\textbf{h}_i} = \mathrm{RNN}(\overrightarrow{\textbf{h}_{i-1}}, \textbf{x}_i).
\end{equation}

In the general NMT model, they also calculate backward hidden vectors of backward RNNs from a reversed source sentence.
However, we only use forward hidden vectors because we cannot use the information of the whole sentence on the simultaneous translation task.

\subsection{Decoder with Attention}
\label{seq:attention_and_decoder}
The decoder takes source hidden vectors as inputs and returns target language words one-by-one with the attention mechanism.
The decoder RNNs recurrently generates target words using its hidden state and an output context.
The conditional generation probability of the target word $\textbf{y}_i$ defined as follows:
\begin{align}
    p_\theta (\textbf{y}_j | \textbf{y}_{<j}, \textbf{x}_{\leq g(j)}) &= \mathrm{softmax}(\textbf{W}_s\tilde{\textbf{b}_j}), \\
    \tilde{\textbf{b}_j} &= \mathrm{tanh}(\textbf{W}_c [ \textbf{c}_j; \textbf{d}_j]), \\
    \textbf{d}_j &= \mathrm{RNN}(\textbf{d}_{j-1}, \textbf{y}_{j-1}).
\end{align}
Here, $\textbf{W}_c, \textbf{W}_p$ are trainable parameters and $\textbf{c}_j$ is a context vector to retrieve source language inputs in forms of a weighted sum of the source hidden vectors $\textbf{h}_j$, defined as follows.
\begin{align}
    \textbf{c}_j &= \sum^{g(j)}_{t=1}{\alpha_{ij} \overrightarrow{\textbf{h}_i}}, \\
    \alpha_{ij} &= \frac{
        \mathrm{exp}(score(\textbf{d}^T_j, \overrightarrow{\textbf{h}_i}))
    }{
        \sum^{g(j)}_{t'=1}{\mathrm{exp}(score(\textbf{d}^T_j, \overrightarrow{\textbf{h}_{t'}}))
    }}.
\end{align}
The \textit{score} function above can be defined in some different ways as discussed by \citet{luong-pham-manning:2015:EMNLP}.
In this paper, we use \textit{dot} attention for this \textit{score} function.

\section{Proposed Method}
In this work, we proposed the method to decide the output timing adaptively.
The proposed method introduces a special token \wait\ which is output instead of delaying translation to target-side vocabulary.

In this section, we first review a standard objective function, softmax cross-entropy and show the problem that occurs when this function is applied to \wait\ (\S\ref{seq:sce}).
After that, we introduce an objective function, called Connectionist Temporal Classification, to handle this problem (\S\ref{seq:ctc}).
Finally, we propose a new objective function to adjust a trade-off between translation accuracy and latency (\S\ref{seq:delay_panalty}) and explain how to combine these objective functions (\S\ref{seq:combine_objective_functions}).

\subsection{Softmax Cross-Entropy}
\label{seq:sce}
Softmax Cross-Entropy (SCE) is a commonly used token-level objective function for multi-class classification including word generation in NMT, defined as follows:
\begin{equation}
    \ell_{ent} = -\sum_{j=1}^{J}{\sum_{k=1}^{K}{\textbf{y}_{jk}\log{p_\theta(\textbf{y}_{jk} | \textbf{y}_{<j}, \textbf{x}_{<g(j)})}}},
\end{equation}
where $\textbf{y}_{ij}$ is a j-th element of the one-hot vector corresponding to the i-th words of the reference sentence and $p(\textbf{y}_{jk}|\cdot)$ is the generation probability of $\textbf{y}_{jk}$.

A correct sequence that corresponds to an output sequence one-by-one is necessary to use SCE as an objective function for NMT.
However, in the proposed method, we cannot simply use SCE because we don't know when we should cause delay.
To avoid this problem, we set the loss for delay tokens to 0 during the time step $t\ (t \leq g(I))$ which the model can output \wait, or while a source sentence is inputted.

\subsection{Connectionist Temporal Classification}
\label{seq:ctc}
As we mentioned in the previous section, we set the loss value for \wait\ to 0, but this causes the problem that it does not optimize about generating \wait .
Against this problem, we use an objective function called Connectionist Temporal Classification (CTC) \cite{graves2006connectionist} for sequence-level optimization.

CTC extends output sequence, called Path $\bm{\pi} = \Omega(\textbf{y})$, to the length $T$ by allowing token repetitions and outputting \wait.
Conversely, we can obtain an original output sequence $\textbf{y} = \Omega^{-1}(\bm{\pi})$ by removing \wait\ and all token repetitions.
The objective function is defined the sum of the probabilities of all possible paths $\bm{\pi} \in \Omega(\textbf{y})$ by using the forward-backward algorithm, as follows:
\begin{align}
    \ell_{ctc}
        &= \sum_{\bm{\pi} \in \Omega (\textbf{y})}{p(\bm{\pi} | X)} \nonumber \\
        &= \sum_{\bm{\pi} \in \Omega (\textbf{y})}{\prod_{t=1}^{T}{p(\pi_t | \pi_{<t}, \textbf{x}_{g(t)})}},
\end{align}
where $\pi_t$ is a t-th element of $\bm{\pi}$.

\subsection{Delay Penalty}
\label{seq:delay_panalty}
Furthermore, we introduce a new objective function, called Delay Penalty, to control latency.
We use this function only when an output token causes the delay; that is, when the model outputs \wait\ or the same token as a previous one.
Delay Penalty is defined by a negative log-likelihood of the probabilities for non-delayed tokens, as follows:
\begin{align}
    \ell_{def} &= -\sum_{t=1}^{T}{\log (1 - w_t)}, \\
    w_t &= \begin{cases}
        p(\wait | \textbf{y}_{<t}, \textbf{x}_{<g(t)}) & (\textbf{y}_t = \wait) \\
        p(\textbf{y}_{t-1} | \textbf{y}_{<t}, \textbf{x}_{<g(t)}) & (\textbf{y}_t = \textbf{y}_{t-1}) \\
        0 & (otherwise)
    \end{cases}
\end{align}

\subsection{Objective Function}
\label{seq:combine_objective_functions}
For optimization, we combine three objective functions introduced so far, as follows:
\begin{equation}
    \ell = \ell_{ent} + \ell_{ctc} + \alpha \ell_{del} .
\end{equation}
Here, $\alpha$ is a hyperparameter to adjust the amount of latency directly.

\section{Experiments}
We conducted simultaneous translation experiments from English to Japanese and discussed accuracy, latency, and issues for translation results.

\subsection{Settings}
All models were implemented as described in the previous sections using PyTorch\footnote{\url{https://pytorch.org}}.
Both the encoders and the decoders were two-layered uni-direcitional LSTM \cite{hochreiter1997lstm}, and the decoder used input feeding\cite{luong-pham-manning:2015:EMNLP}.
The number of dimensions in word embeddings and hidden vectors was set to 512, and the minibatch size was 64.
We use Adam \cite{Kingma2015adam} for optimization with the default parameters.
The learning rate was set to $10^{-1}$, and gradient clipping was set to 5. 
The dropout probability was set to $0.3$.
The learning rate was adjusted by a decay factor of $1/\sqrt{2}$ when the validation loss was larger than that in the previous epoch.
Then, we chose the best parameter/model with the smallest validation loss for evaluation.

We used two different corpora for the experiments: small\_parallel\_enja\footnote{\url{https://github.com/odashi/small_parallel_enja}} and Asian Scientific Paper Excerpt Corpus (ASPEC) \cite{NAKAZAWA16.621}.
small\_parallel\_enja is a small-scale corpus that is consist of sentences filtered sentence length 4 to 16 words, and ASPEC is a mid-scale corpus of the scientific paper domain.
Table \ref{table:corpora} shows their detailed statistics.
\begin{table}[tbp]
    \centering
    \caption{Number of sentences for each corpus used in the experiments.}
    \label{table:corpora}
    \begin{tabular}{c||c|c|c}\hline
        \multirow{2}{*}{Corpus} & \multicolumn{3}{c}{Number of Sentence} \\ \cline{2-4}
        & Train & Valid. & Test \\ \hline\hline
        small\_parallel\_enja & 50k & 500 & 500 \\ \hline
        ASPEC & 964k & 1790 & 1812 \\ \hline
    \end{tabular}
\end{table}

All datasets were tokenized into subword unit \cite{sennrich-haddow-birch:2016:P16-12:BPE,kudo-richardson-2018-sentencepiece} by using Sentencepiece \footnote{\url{https://github.com/google/sentencepiece}}.
The source and target language vocabularies were independent, and their size was set to 4000 tokens for small\_parallel\_enja and 8000 tokens for ASPEC, respectively.
We filtered out the sentence whose number of tokens was more than 60 tokens, or the length ratio was more than 9 from the training set.

We used ``{\bf Wait-k}'' models and general NMT models as baseline models.
General NMT models were attention-based encoder-decoder and it translated sentences from full-length source sentences (called {\bf Full Sentence}).
For evaluation metrics, we used BLEU \cite{papineni2002bleu} and RIBES \cite{isozaki-etal-2010-automatic} to measure translation accuracy, and token-level delay to measure latency.
We used Kytea \cite{neubig-etal-2011-kytea} as a tokenize method for evaluations of Japanese translation accuracy.

\subsection{Experiments with Small-scale Corpus}
We conducted small-scale experiments using small\_parallel\_enja.
We compared different hyperparameters: $k = \{3, 5\}$ and $\alpha = \{0, 0.01, 0.03, 0.05\}$.

\begin{table*}[htb]
    \centering
    \caption{Latency and automatic evaluation scores with small\_parallel\_enja. Latencies are shown by averages and standard deviations (in parentheses) in the number of tokens.}
    \begin{tabular}{cc|c||c|c}\hline
        \multicolumn{2}{c|}{Method} & Latency (\#tokens) & BLEU & RIBES \\\hline\hline
        \multicolumn{2}{c|}{Full sentence} & 9.75 ($\pm$ 2.69) & 34.53 & 84.03 \\ \hline\hline
        Wait-k \cite{ma2018stacl} & k=3                      & 3.00 ($\pm$ 0.00) & 31.06 & 82.46 \\
                                  & k=5                      & 5.00 ($\pm$ 0.00) & 33.29 & 83.45 \\\hline\hline
        Ours                      & $\alpha$=0.00            & 4.32 ($\pm$ 3.14) & 28.01 & 81.78 \\
                                  & $\alpha$=0.01            & 4.29 ($\pm$ 3.16) & 30.42 & 82.60 \\
                                  & $\alpha$=0.03            & 2.88 ($\pm$ 2.95) & 26.47 & 80.51 \\
                                  & $\alpha$=0.05            & 0.80 ($\pm$ 1.96) & 22.60 & 77.86 \\\hline
    \end{tabular}
    \label{table:exp1_result}
\end{table*}
Table~\ref{table:exp1_result} shows the results in latency and automatic evaluation scores on small\_parallel\_enja.
The full sentence scores are upper bounds of incremental methods.
The proposed method reduced the average latency in more than 50\% from the full sentence baseline
with some loss in BLEU and RIBES.
The BLEU and RIBES results by the proposed method were worse than those by Wait-k.
Th would be due to some degradation in smaller latency parts that were determined adaptively by the proposed methods
while Wait-k keeps the fixed latency.

\subsection{Experiments with Mid-scale Corpus}
We investigated the performance on longer and more complex sentences by the experiments using ASPEC.
We compared different hyperparameters: $k = \{5, 7\}$ and $\alpha = \{0.03, 0.05, 0.1\}$.

\begin{table*}[htb]
    \centering
    \caption{Latency and automatic evaluation scores with ASPEC.}
    \label{table:exp2_result}
    \begin{tabular}{cc|c||c|c}\hline
        \multicolumn{2}{c|}{Method} & Latency (\#tokens) & BLEU & RIBES \\\hline\hline
        \multicolumn{2}{c|}{Full sentence} & 29.81 ($\pm$ 14.30) & 32.22 & 80.17 \\ \hline\hline
        Wait-k \cite{ma2018stacl} & k=5                      &  5.00 ($\pm$  0.00) & 21.53 & 71.40 \\
                                  & k=7                      &  7.00 ($\pm$  0.00) & 23.20 & 73.21 \\ \hline\hline
        Ours                      & $\alpha$=0.03            & 23.03 ($\pm$ 14.08) & 24.86 & 72.59 \\
                                  & $\alpha$=0.05            & 21.96 ($\pm$ 13.88) & 22.45 & 70.60 \\
                                  & $\alpha$=0.1             & 17.13 ($\pm$ 12.69) & 23.66 & 72.27 \\ \hline
    \end{tabular}
\end{table*}
Table~\ref{table:exp2_result} shows the results in latency and automatic evaluation scores on ASPEC.
We can see the proposed method showed much larger latency than Wait-k.
This is probably due to many long and complex phrases used in scientific articles in ASPEC.
Wait-k has to translate such a long phrase without sufficient input observations due to its strict fixed latency strategy.
On the other hand, the proposed method can wait for more input tokens adaptively by generating \wait
at the cost of large latency.

\subsection{Discussion}
In the experimental results above,
the proposed method determined the translation latency adaptively,
short delay for short and simple inputs as in small\_parallel\_enja and
long delay for long and complex inputs as in ASPEC.
Here we discuss our results in detail using some examples.

Table~\ref{table:example} shows translation examples on small\_parallel\_enja.
In the first example, the proposed method gives a correct translation result by adaptive waits.
Wait-k generated unrelated words \cjk{野球} (\textit{baseball}) and \cjk{飲-み} (\textit{drink}) due to the poor input observations with its small fixed latency.
The proposed method waited until a subword \textit{swim} was observed and successfully generate a word \cjk{泳-ぐ} (\textit{swim}).
\begin{table*}[htp]
    \centering
    \caption{Translation examples in small\_parallel\_enja. \waits shows the generation of \wait token.}
    \label{table:example}
    \includegraphics[width=\linewidth]{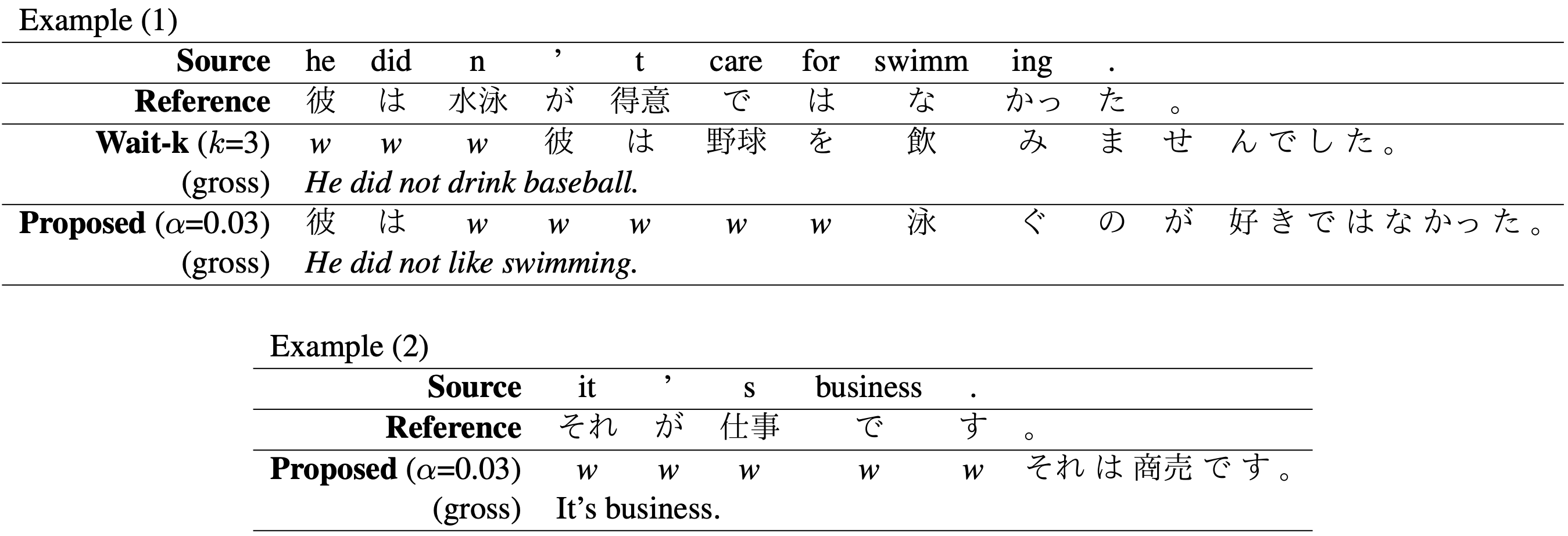}
\end{table*}

However, the proposed method sometimes generated consecutive \wait symbols until the end of input, as shown in the second example.
This is probably due to our training strategy;
the latency penalty would not be large enough to choose small latency translation
at the cost of some increase in SCE- and CTC-based loss.
The translation data in the experiments are not from simultaneous interpretation but standard translation,
so the current task does not match with the proposed approach.
The use of specialized data for simultaneous translation would be important in practice,
such as \textit{monotonic translations} like simultaneous translation.

\section{Conclusion}
In this paper, we proposed an adaptive latency control method for simultaneous neural machine translation
in syntactically distant language pairs.
We introduced a meta token \wait to wait until the observation of the next input token.
We proposed a CTC-based loss function to perform optimization using bilingual data without appropriate positions of \wait,
which is used along with the latency penalty and a standard word prediction loss.
The experimental results suggest the proposed method determines when to translate or when to wait in an adaptive manner.
Future work includes further analyses on translation accuracy in different latency conditions and time-based latency evaluation instead of the token-based one.

\section*{Acknowledgments}
A part of this work is supported by JSPS Kakenhi JP17H06101.

\bibliography{reference}
\bibliographystyle{acl_natbib}

\end{document}